\documentclass[conference]{IEEEtran}

\newcommand{\ignore}[1]{}
\usepackage[pass]{geometry}
\usepackage{fancyhdr}
\usepackage[normalem]{ulem}
\usepackage[hyphens]{url}
\usepackage{hyperref}

\usepackage{booktabs}
\usepackage{multirow}
\usepackage{graphics}
\usepackage{algpseudocode}
\usepackage{pseudocode}
\usepackage{algorithm}
\usepackage{comment}
\usepackage{amsmath,amssymb}
\usepackage{latexsym}
\usepackage{graphicx}
\usepackage{subfigure}
\usepackage{epstopdf}
\usepackage{epsfig}
\usepackage{xspace}
\usepackage[bottom]{footmisc}
\usepackage{amsmath}

\usepackage{mathptmx}
\usepackage[11pt]{moresize}
\usepackage{todonotes}
\usepackage{array}

\usepackage{mathptmx} 
\usepackage{fancyhdr}
\usepackage[normalem]{ulem}
\usepackage[hyphens]{url}
\usepackage[sort,compress]{cite}
\usepackage[final]{microtype}
\usepackage{flushend}

\newcolumntype{L}[1]{>{\raggedright\let\newline\\\arraybackslash\hspace{0pt}}m{#1}}
\newcommand{\Design}{RAPIDNN\xspace}
\newcommand{\RNA}{RNA\textsc{}\xspace}
\newcommand{\ReCA}{NDCAM\xspace}
\renewcommand{\baselinestretch}{0.9}

\newcommand{\squeezeup}{\vspace{-1mm}}
\newcommand{\squeezeupless}{\vspace{-0.1mm}}

\graphicspath{ {images/} }

\usepackage{color}

\ifx{
\fancypagestyle{firstpage}{
  \fancyhf{}
  
  \fancyhead[C]{\vspace{15pt}\normalsize{MICRO 2018 Submission
      \textbf{\#\microsubmissionnumber} -- Confidential Draft -- Do NOT Distribute!!}} 
  \fancyfoot[C]{\thepage}
}
}\fi

\begin{document}
\title{\Design: In-Memory Deep Neural Network Acceleration Framework}
\author{
    Mohsen Imani$^{*}$, Mohammad Samragh$^{\ddag}$, Yeseong Kim$^{*}$, Saransh Gupta$^{*}$,\\ Farinaz Koushanfar$^{\ddag}$ and Tajana Rosing$^{*}$\\
        $^{*}$Computer Science and Engineering Department, UC San Diego, La Jolla, CA 92093, USA\\
    $^{\ddag}$Department of Electrical and Computer Engineering, UC San Diego, La Jolla, CA 92093, USA\\
}
\maketitle

\begin{abstract}
Deep neural networks (DNN) have demonstrated effectiveness for various applications such as image processing, video segmentation, and speech recognition.
Running state-of-the-art DNNs on current systems mostly relies on either general-purpose processors, ASIC designs, or FPGA accelerators, all of which suffer from data movements due to the limited on-chip memory and data transfer bandwidth.
In this work, we propose a novel framework, called \Design, which
performs neuron-to-memory transformation in order to accelerate DNNs in a highly parallel architecture. 
\Design reinterprets a DNN model and maps it into a specialized accelerator, which is designed using non-volatile memory blocks that model four fundamental DNN operations, i.e., multiplication, addition, activation functions, and pooling.
The framework extracts representative operands of a DNN model, e.g., weights and input values, using clustering methods to optimize the model for in-memory processing. 
Then, it maps the extracted operands and their pre-computed results into the accelerator memory blocks.
At runtime, the accelerator identifies computation results based on efficient in-memory search capability which also provides tunability of approximation to improve computation efficiency further.
Our evaluation shows that \Design achieves 68.4$\times$, 49.5$\times$ energy efficiency improvement and 48.1$\times$, 10.9$\times$ speedup as compared to ISAAC and PipeLayer, the state-of-the-art DNN accelerators, while ensuring less than 0.5\% quality loss.

\end{abstract}

\section{Introduction}
\label{sec:intro}
The emergence of \textit{Internet of Things} (IoT) significantly increases sizes of application datasets required to be processed~\cite{atzori2010internet}.
As a solution which automatically extracts useful information from the largely generated data, artificial neural networks have been actively investigated. 
In particular, deep neural networks (DNNs) demonstrate superior effectiveness for diverse classification problems, image processing, video segmentation, speech recognition, computer vision, and gaming~\cite{oquab2014learning, lecun2010convolutional, ji20133d, clark2014teaching}.
Although many DNN models are implemented on high-performance computing architectures such as GPGPUs by parallelizing tasks, 
running neural networks on the general purpose processors is still slow, energy-hungry, and prohibitively expensive~\cite{krizhevsky2012imagenet}.

Earlier work proposed FPGAs~\cite{gupta2015deep, sharma2016high, zhang2015optimizing, ma2017optimizing, nazemi2018nullanet} and ASIC  designs~\cite{luo2017dadiannao,reagen2016minerva,chen2014diannao, chen2014dadiannao, hegde2018ucnn} to accelerate neural networks. However, these techniques pose a critical technical challenge due to data movement cost, since they require dedicated memory blocks, e.g., SRAM, to store the large size of network weights and input signals. In the context of efficient DNN implementation, 
prior works employ a variety of techniques to optimize the enormous computation cost, yet the memory still takes up to 90\% of the total energy consumption to perform DNN inference tasks even in highly optimized ASIC designs~\cite{reagen2016minerva}. 

Processing in-memory (PIM) is a promising solution to address the data movement issue by implementing logics within a memory~\cite{gokhale1995processing, ahn2015pim, ahn2015scalable, li2016pinatubo, boroumand2017lazypim}.
Instead of sending a large amount of data to the processing cores for computation, PIM performs a part of computation tasks, e.g., bit-wise computations, inside the memory; thus the application performance can be accelerated significantly by avoiding the memory access bottleneck. 
Several existing works have proposed PIM-based neural network accelerators which keep the input data and trained weights inside memory~\cite{chi2016prime, shafiee2016isaac}. 
For example, the work in~\cite{shafiee2016isaac} showed that memristor devices could model the input-weight multiplications of each neuron in a crossbar memory. These approaches store the trained DNN weights as device resistance values, and then pass input values as an analog voltage to these devices~\cite{serrano2013stdp}. 
Although these approaches are the first pace towards employing PIM for DNN acceleration, they have three significant downsides:
(i) They utilize Analog to Digital Converters (ADCs) and Digital to Analog Converters (DACs) which take the majority of the chip area and power consumption, e.g., 89\% of chip power in~\cite{shafiee2016isaac}. In addition, the mixed-signal ADC/DAC blocks do not scale as fast as the memory device technology does.  
(ii) The existing PIM approaches use multi-level memristor devices that are not sufficiently reliable for commercialization unlike commonly-used single-level NVMs, e.g., Intel 3D Xpoint~\cite{3Dxpoint}. 
(iii) Finally, they only support matrix multiplication in analog memory while other operations such as activation functions are implemented using CMOS-based digital logic. This makes the design non-generic and increases the expense of fabrication.

In this paper, we propose a novel DNN acceleration framework, called \Design, which performs neuron-to-memory transformation to accelerate DNN in a highly parallel architecture. \Design supports \textit{all} DNN functionalities in a \textit{digital-based} memory design. 
\Design first analyzes computation flows of a DNN model and encodes key DNN operations for a specialized PIM-enabled accelerator.
Our framework identifies representative parameters processed in each neuron, i.e., weights and input values, using clustering algorithms.
The other key operations, e.g., activation functions, are also approximately modeled to enable in-memory processing.
Based on these techniques, we create a new DNN model which is compatible with the memory-based accelerator.

The key finding underlying this procedure is that, even though the operations of a DNN model, e.g., multiplications and activation functions, are continuous functions, they can be approximated as step-wise functions without losing the quality of inference. Once a step-wise approximation is developed, we can create computation tables which store the finite pre-computed values, and map them into specialized memory blocks capable of in-memory computations. The naive solution for step-wise approximation would employ linear quantization to represent the inputs (operands) and outputs of pertinent functions~\cite{hubara2016quantized}. To ensure maximum accuracy of the step-wise approximation, we propose to employ a non-linear quantization which takes account of statistical properties of each operand and output within the DNN, thus improving the accuracy. For example, although we quantize the Rectified Linear Unit (ReLU) activation function with 64 pairs for inputs and outputs, the inference accuracy can be maintained at the same level.

The proposed \Design framework supports three layers popularly used for designing a DNN model: fully-connected, convolution, and pooling layer.
We group the computation tasks of the networks by four operations, multiplication, addition, activation function, and pooling.
Our accelerator supports the multiplication and addition operations inside a crossbar memory, and other operations, activation function and pooling, are modeled with associative memory (AM) blocks which are a form of a lookup table.
The main contribution of the paper is listed as follows: 
\begin{itemize}
\squeezeupless
    \item To the best of our knowledge, \Design is the first neural network accelerator which maps all functionalities inside the memory block. Using direct digital-based computation without any analog-to-digital conversion ensures a scalable design approach for our accelerator.
    In addition, we remove the necessity of using unreliable multi-level memristors by implementing \Design using commonly used single-level memristor devices. 
    \squeezeupless
    \item We present software support for \Design along with novel algorithms which reinterpret DNN models to enable in-memory processing with minimal accuracy loss of DNN inference.
    \item Providing adjustable DNN reinterpretation mechanisms that allow users to configure \Design for different DNN applications optimally.
     We explore how different memory sizes impact the inference accuracy. 
     \squeezeupless
     \item Proof-of-concept evaluations on six DNN applications demonstrate that using small-sized memory blocks, e.g., around 5 KBytes for each neuron, \Design can provide the same level of the prediction quality. For instance, we achieve 68.4$\times$, 49.5$\times$ energy efficiency improvement and 48.1$\times$, 10.9$\times$ speedup on average as compared to ISAAC~\cite{shafiee2016isaac} and PipeLayer~\cite{song2017pipelayer} (state-of-the-art PIM-based DNN accelerators), respectively, while ensuring less than 0.5\% of quality loss.
\end{itemize}

\section{\Design Design}
\subsection{Overview of \Design}
\begin{figure}[t!]
  \centering
   \epsfig{file=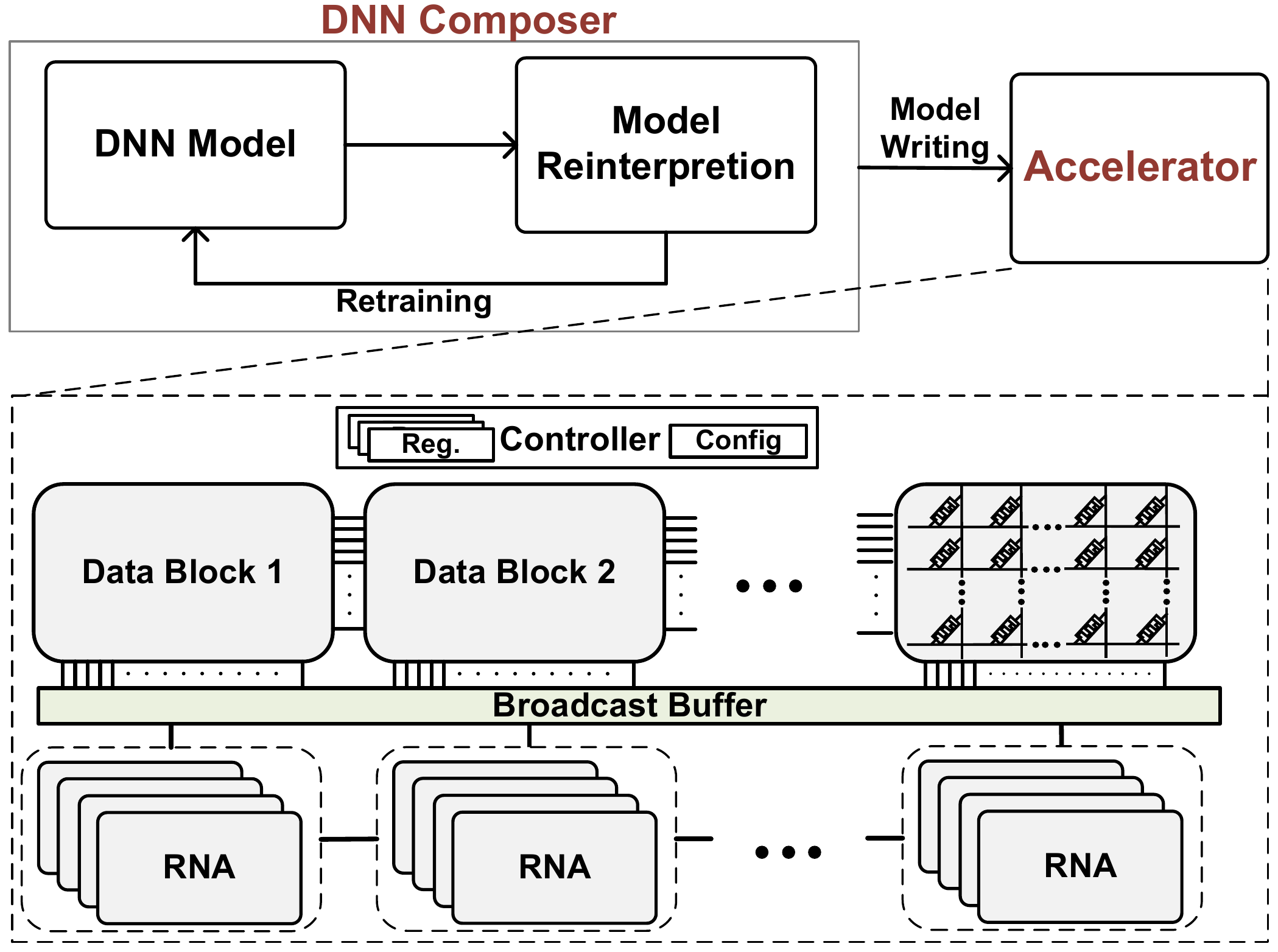, width=0.9\columnwidth}
\squeezeup
\squeezeupless
  \caption{Overview of \Design framework.}
  \squeezeup
  \squeezeup
  \label{fig:Overview}
\end{figure}

Figure~\ref{fig:Overview} illustrates a high-level overview of the proposed \Design framework.
It consists of two interconnected blocks: a software module, \textit{DNN composer} and a hardware module, \textit{accelerator}.
The role of the DNN composer is to convert each neural network operation to a table that can be stored in the accelerator memory blocks which process all neural network computations inside memory. The entries of these tables are operands (inputs) and outputs of pertinent operations, e.g., multiplication and activation functions, that are employed to construct neural networks. We adopt the idea of step-wise function approximation to form input-output tables that can replace CMOS-based logic units of current processors. By statistically analyzing the input and outputs of the corresponding functions in an offline stage, starting with a given \textit{DNN model}, the DNN composer analyzes weights and inputs of each neuron and generates a new DNN model which is compatible to the proposed PIM-based accelerator. Particularly, the output of the DNN composer module is a neural network whose operations can be efficiently implemented using finite tables inside the memory.
The newly constructed DNN model is repeatedly revised through multiple retraining procedures.
After generating the final model through the iterations, it is stored into the accelerator so that it can perform the online inference.

The proposed \Design accelerator supports both memory and computing functionalities by using two different memories, \textit{data} blocks and \textit{\RNA} blocks.
The data block is a typical crossbar memory which stores an input dataset processed by the DNN model.
The resistive neural acceleration (\RNA) blocks designed with multiple memory banks are in charge of processing the DNN.
In the execution phase, each input data is applied to all \RNA blocks in parallel using a memory buffer which keeps them in a FIFO. 
Then, the \RNA blocks, which are the main cores of the \Design accelerator, process the sequence of the input data.
A single \RNA block computes the output for one neuron using multiple internal memories which model the fundamental neural network operations, i.e., multiplication, activation function, and pooling.
Once the inference is completed, the accelerator writes the computed results back to the crossbar memory.
In the next few sections, we describe our strategies to map the DNN to the \Design accelerator.

\begin{figure*}[t!]
\squeezeup
\squeezeup
  \centering
   \epsfig{file=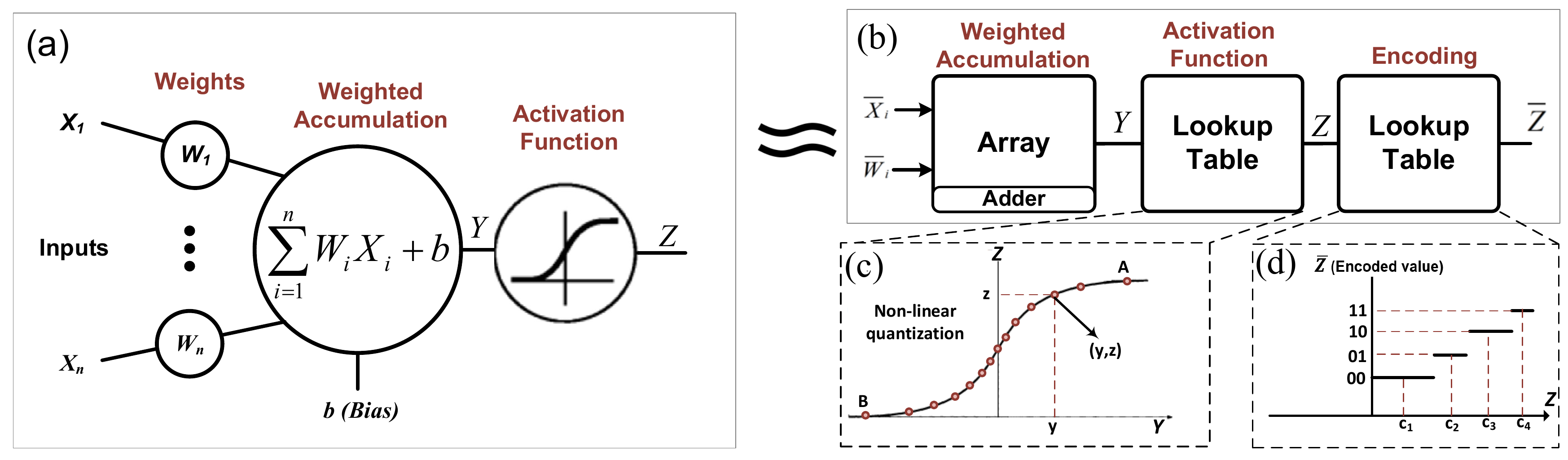, width=0.8\textwidth}
   \squeezeup
   \squeezeupless
  \caption{A representation of computations of a neuron and its reinterpretation in the proposed framework.}
  \squeezeup
  \squeezeupless
  \label{fig:Neuron}
\end{figure*}

\subsection{Preliminary of DNN Reinterpretation}
\label{sec:pre}
A DNN model consists of multiple layers which have multiple neurons. These layers are stacked on top of each other in a hierarchical formation; that is, the output of each layer is forwarded to the next layer. The outputs of the last layer are used for inference. 
In this paper, we focus on three types of layers that are most commonly utilized in designing efficient neural networks: (i) convolution layers, (ii) fully connected layers, and (iii) pooling layers. \Design is inherently capable of applying pooling layers without any modification of the neural network. For convolution and fully connected layers, the framework reinterprets the layers in an offline process to ensure compatibility with the memory-based accelerator.

Figure~\ref{fig:Neuron}a depicts one neuron which computes its output in two steps: (i) weighted sum and (ii) activation function computation.
The neuron takes a vector of neuron values from the preceding layer $\mathbf{X}=\langle X_0, \cdots, X_n \rangle$, then computes its output as follows $\varphi(\sum_{i=1}^{n}{W_iX_i}+b)$, where $W_i$ and $X_i$ correspond to a weight and an input respectively, $b$ is a bias parameter, and $\varphi(.)$ is a nonlinear activation function. 

In the \Design framework, we interpret the computations of a neuron to a series of operations shown in Figure~\ref{fig:Neuron}b to make the DNN compatible with the proposed accelerator. We describe each operation below in details.

\noindent{\bf Weighted accumulation:} 
There are two basic operations required for weighted accumulation: multiplication and addition. Here we consider the multiplication operation, while we address additions in Section~\ref{sec:in_mem-add}. Consider the two operands of a multiplication, $a$ and $b$, where each operand belongs to a finite set. For instance, in a 32-bit floating-point representation, each input can take one of $2^{32}$ different possibilities. If we could store all pairwise multiplications (i.e., $2^{32}\times 2^{32} = 2^{64}$ possibilities) in an array beforehand, we could fetch the correct result from the array instead of performing actual multiplication using CMOS logic. Obviously, in this naive approach, the size of pairwise results would be unacceptably huge to create an array in real-world systems.
Thus, the key technical challenge is how to reduce the size of two input sets.

We propose to reduce the input span by carefully selecting a subset from the input spaces, called ``best representatives,'' and approximating every input operand by its closest representative.
In our design, the DNN composer selects the best representatives by analyzing the weights and input values given to the networks (Section~\ref{sec:clustering}). For instance, we may find $4$ values to account for each input operand, in which case we would have $4\times 4=16$ different possible output values. In practice, our experiments show that using a maximum number of 64 representatives (4096 possible outputs) can fully recover the DNN accuracy. 

\begin{figure}
\centering
\includegraphics[width=1\columnwidth]{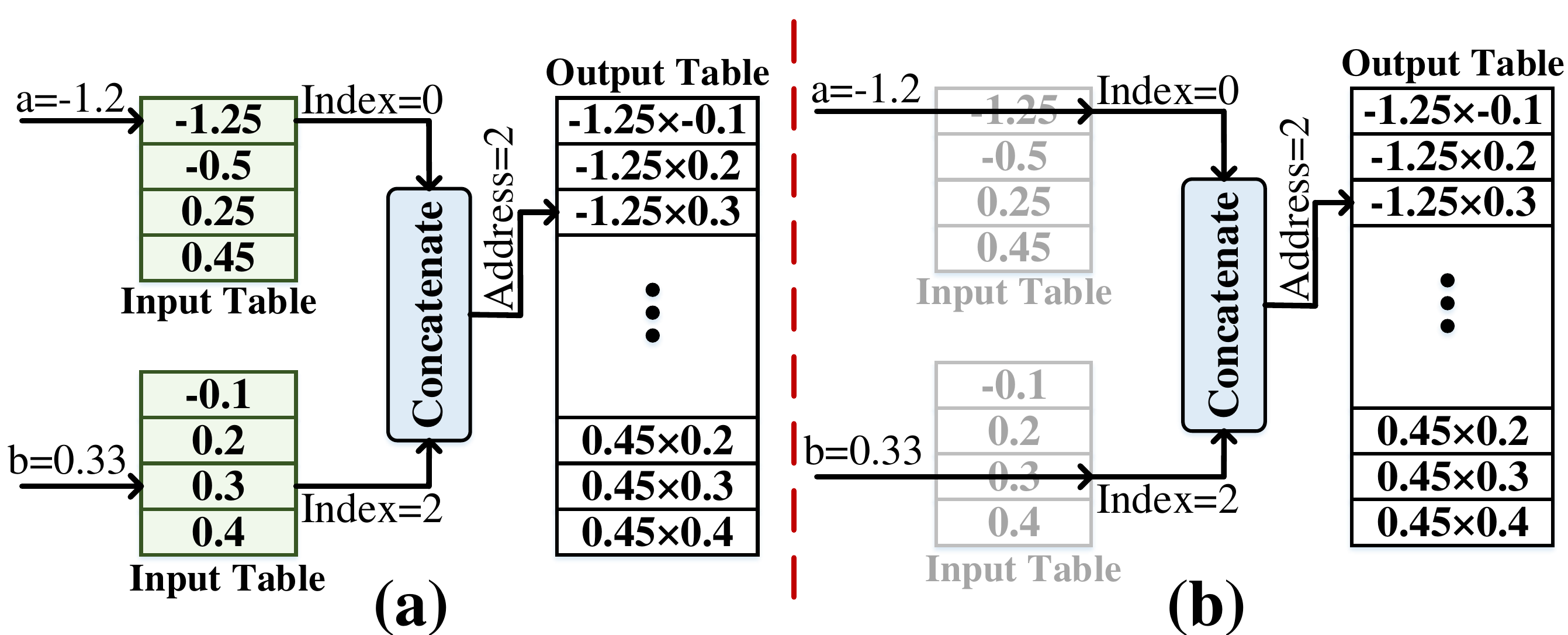}
\squeezeup
\squeezeup
\caption{(a) In-memory approximate multiplication using raw data. (b) Multiplication using encoded data.}
\squeezeup
\squeezeup
\squeezeupless
\label{fig:mult}
\end{figure}

Figure~\ref{fig:mult}a presents the schematic view of an example memory based multiplier which is configured to operate using 4 representatives. For each operand, the first step is to determine which entry in the table is the closest value. Each input table generates an index to the corresponding closest representative. Therefore, the approximate multiplication result can be fetched from the output table according to the indices generated by the two input tables. This design requires two lookup tables for the input operands; however, below we describe how we can completely remove the input tables and simply replace them with wires.

Note that the operands and the outputs can be mapped into the set of best representatives using fewer bits, e.g., 2-bits for inputs ($2^2=4$ possibilities) based on one-to-one correspondences. We call elements of the mapped set as \textit{encoded values}. In particular, for every weight value $W_i$ and neuron value $X_i$, we denote the encoded values by $\overline{W}_i$ and $\overline{X}_i$. Figure~\ref{fig:mult}b shows how encoded operands can facilitate the in-memory multiplication: there is no need to search for the closest value in the input tables as the inputs themselves represent the indices; thus, the input tables can simply be replaced by wires. The first operand ($\overline{W}_i$) is simply encoded offline and stored in the weight matrix. The second operand ($\overline{X}_i$) is encoded during DNN execution after the neuron output is computed in the preceding layer.

\noindent{\bf Activation function:}
We also model the activation function for enabling PIM.
Neural networks use different types of activation functions.
For example, ``sigmoid'' has been used as one of the basic activation functions~\cite{hansen1990neural}, and there are other activation functions which recently gain popularity due to the better inference accuracy for some applications, e.g., ''Rectified linear unit'' (ReLU) and ''Softsign'' ~\cite{nair2010rectified, glorot2010understanding}.
One way to support different activation functions is to exploit different CMOS-based logic, but they may be expensive to fabricate and could not support other activation functions.
In our design, we approximately model an activation function using a small lookup table.
Using this approach, we can represent any activation function.
Figure~\ref{fig:Neuron}c shows this procedure for the sigmoid function as an example.
A lookup table stores multiple $(y, z)$ coordinates of the activation function.
For a given input value, (i.e., the output of the weighted accumulation $Y$), the table identifies a stored coordinate whose value is closest to the input and generates the corresponding output $z$. We elaborate on the definition of ``closeness'' and the hardware implementation of the table in Section~\ref{sec:NNCAM}. 

Since a typical activation function is saturated for either very large or small input values, we can effectively limit the domain using two upper and lower points ($A$ and $B$ in Figure~\ref{fig:Neuron}) with a minimal quality change.
We can equally or non-equally quantize the range from $A$ and $B$ to select the intermediate values. Intuitively, the accuracy of the approximated function mainly depends on the number of values in the lookup table.
For example, increasing the number of data points provides better accuracy.
Non-linear quantization enables putting more points on the regions that activation function has sharper changes. This way of quantization improves the quality of approximation.  
Note that the proposed technique ensures the generality of the algorithm. However, for easy activation functions such as ReLU, our design can replace the lookup table with a simple comparator block.  

\label{sec:support}

\begin{figure*}[t!]
\squeezeup
\squeezeup
\squeezeupless
  \centering
   \epsfig{file=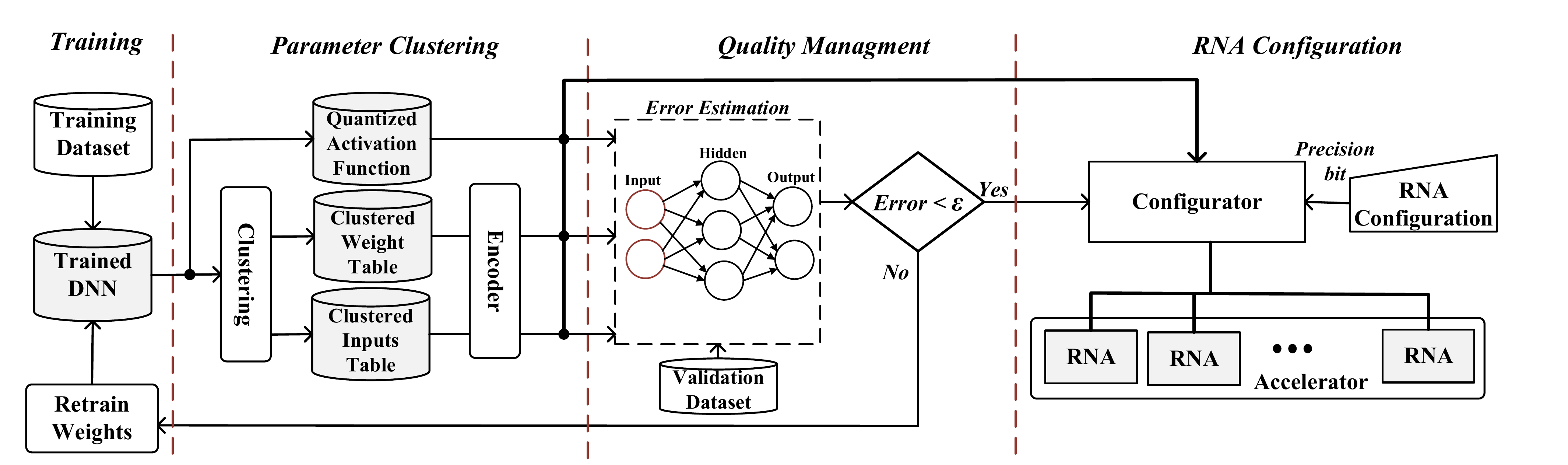, width=0.8\textwidth}
\squeezeup
  \caption{DNN composer which reinterprets a model and configures the proposed accelerator.}
  \squeezeup
  \label{fig:framework}
\end{figure*}

\noindent{\bf Encoding block:}
Since the neurons of our reinterpreted model operate on \textit{encoded} values, we need to convert the output of the activation function into an encoded value.
For this purpose, we utilize a lookup table with a similar structure to the one used for activation function modeling.
Figure~\ref{fig:Neuron}d presents an example of encoding into 2-bits (4 representatives).
Since the encoded value for the activation units, $\overline{Z}$, is used as the input of the neurons of the next layer, say $\overline{X}_j$, we encode the outputs based on their similarity to the representatives corresponding to the next DNN layer. 
In the case of the input layer, to encode each raw input data, we add one more \textit{virtual} layer as an initial layer of the DNN.
The neuron of this layer does not perform any computation tasks, i.e., the weighted accumulation and an activation function, but only encodes the input values to pass them to the first computation layer, e.g., fully connected or convolution layer.

\section{DNN Composer}
Figure~\ref{fig:framework} shows the overall procedure of the DNN composer.
The DNN composer performs the DNN reinterpretation in an offline stage in four main steps: parameter clustering, quality management, network retraining, and \RNA configuring.

The parameter clustering module uses the pre-trained DNN model and the training data to find the best representatives for each layer's inputs and weights. In particular, we use the k-means algorithm~\cite{lloyd1982least} and interpret the resulting centers of clusters as the representative values. Once the multiplication, activation function, and encoding tables are generated for each DNN layer, the error estimation module evaluates the reinterpreted memory-based DNN on the validation data.
If an error criterion is not satisfied, the model is retrained under the modified condition, so that the model is more fitted with the clustered weights. We proceed the same procedure until an error rate, $\epsilon$, is satisfied or a pre-defined number of iterations is repeated. After the iterations, the new model compatible with the proposed accelerator is stored into the accelerator for real-time inference.

\subsection{Multiplication Operand Clustering}\label{sec:clustering}
As discussed in Section~\ref{sec:pre}, the proposed \Design framework converts key arithmetic computations to memory-based computations to reduce the cost of data movement.
The first key procedure is to identify the best representatives for multiplication based on $k$-means clustering.
Assuming that the actual numerical values belong to the set $\theta$, the objective of the clustering algorithm is to find a set of $k$ cluster centroids $\{c_1,\ c_2,\ \dots,\ c_k\}$ that can best represent the values within $\theta$.
Formally, the objective is to reduce the Within Cluster Sum of Squares (WCSS):
\squeezeup
\begin {equation}\label{eq:Kmeans}
        \min_{c_{1},\ c_2,\ \dots,c_{k}} \ (WCSS=\sum_{j=1}^{k}{\sum_{\theta_i\in c_{j}}||\theta_i-c_{j}||^2})
        \vspace{-0.3cm}
\end{equation}
where $\theta_i$ is the $i^{th}$ sample drawn from $\theta$ and $k$ is the number of clusters.
In the rest of this paper, we refer to the set of these representatives found in the clustering procedure as a \textit{codebook}.
We use the $k$-means clustering algorithm to solve the minimization objective for each neural network layer separately, as the distribution of weights and inputs can vary across different layers. The weights and inputs are clustered differently as follows:

\begin{itemize}
    \item {\bf Weights}: The weights of each layer are fixed in the inference phase; therefore, to form the codebook for the fixed parameters, the clustering algorithm is applied on the fixed weights. Assuming that a fully-connected layer maps $N$ neurons into $M$ outputs, the corresponding matrix $W_{M\times N}$ is clustered once, and a single codebook is generated for the whole matrix. For convolution layers, the weights corresponding to different output channels are clustered separately: a convolution layer mapping $N$ channels into $M$ channels using a weight tensor $W_{h\times h\times N \times M}$ is divided into $M$ different tensors and each tensor is clustered separately, resulting in $M$ different codebooks. 
    \item{\bf Inputs}: The input of each layer is determined by its preceding layer, hence, the inputs of all layers depend on the raw data given to the network; therefore, we execute the feed-forward procedure with the training dataset to form $\theta$ for each DNN layer, then apply $k$-means on this $\theta$ to find the corresponding codebook.
    In our implementation, we run the network with a set of inputs randomly sampled from the training dataset.
    The sampling technique significantly reduces the overhead of computing the codebook as our experiments show that sampling as low as $2\%$ of the data is sufficient to achieve reasonable accuracy.
\end{itemize}

\noindent{\bf Multi-level clustering:} The codebook size determines the multiplications precision with the lookup table-based approach: the more cluster centroids are chosen, the more the precision will be. Note that this is the numerical precision and the classification accuracy (the objective of the neural network) depends on the application too.
Some applications would require more fine-grained clusters in order to deliver reasonable classification accuracy, while other applications might show high classification accuracy with smaller numerical precision.

\begin{figure}
    \centering
    \includegraphics[width=0.8\columnwidth]{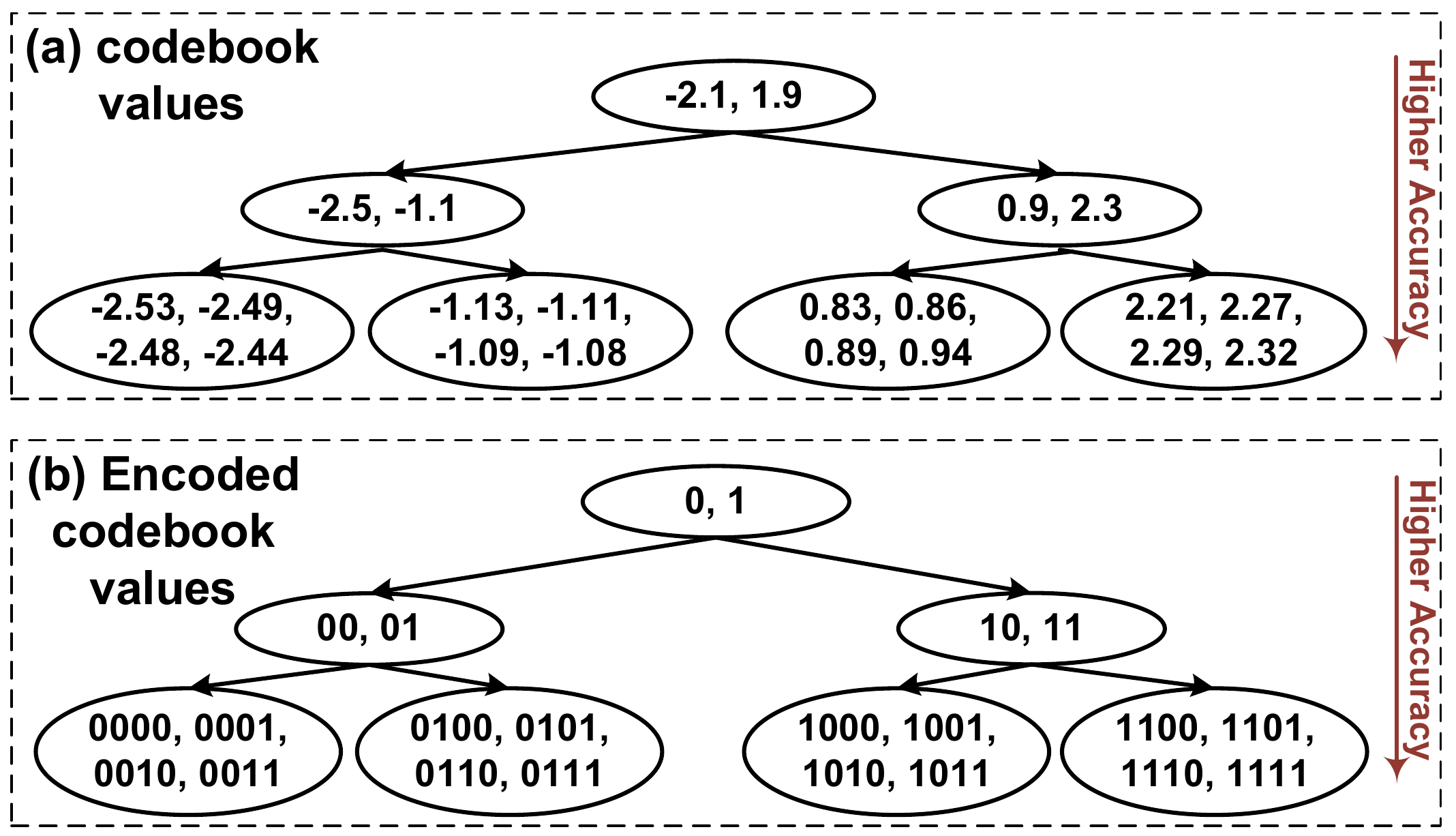}
    \squeezeup
    \squeezeupless
    \caption{Illustration of tree-based codebook generation.}
    \squeezeup
    \squeezeupless
    \label{fig:Hcluster}
\end{figure}

To offer flexibility for configuring the accelerator, we propose a multi-level clustering method which creates the codebook as a tree. Figure~\ref{fig:Hcluster}a shows an example of the tree-based codebook. The first level includes 2 cluster centroids: $\{-2.1,\ 1.9\}$; in the second level, each cluster is again partitioned into 2 separate clusters that more accurately represent the data.
For instance, the cluster representing $1.9$ in the first level is partitioned into $\{0.9,\ 2.3\}$ in the second level to provide more precision.

The tree is created by recursively calling the $k$-means clustering module. First, the $k$-means module clusters the whole $\theta$ into two clusters: $\theta_1$ and $\theta_2$ represented by codebook values -2.1 and 1.9, respectively. Next, $\theta_1$ and $\theta_2$ are separately partitioned to two different clusters, so that each sub-cluster itself is represented using a codebook of 2 values. This recursive process is continued to create the last level of the tree (three levels in this example), and then all codebook values are computed.

Figure~\ref{fig:Hcluster}b shows the encoding tree for the same hierarchical codebook. Deeper layers' encodings are formed by appending extra bits to those of their parent nodes in the tree. Deeper levels provide higher multiplication precision, whereas shallower levels deliver less precision but reduce the area overhead and power consumption. As such, the accuracy can be dynamically tuned for different applications.
Note that the codebook values in each level are sorted before encoding; thus, comparison over the encoded values has the same output as a comparison over the original codebook values.
This property enables \RNA to perform max-pooling over the encoded data. We explain how the hardware accelerator implements the pooling functionality in Section~\ref{sec:actsenpool}.

\subsection{Quality Estimation and Model Retraining}
\label{sec:qual}
We retrain the model with the reinterpreted condition to ensure better accuracy. This procedure is done by two steps, weight retraining and error estimation described below.

\begin{figure} [t]
\squeezeup
\centering
\subfigure
{\includegraphics[width=1.6in]{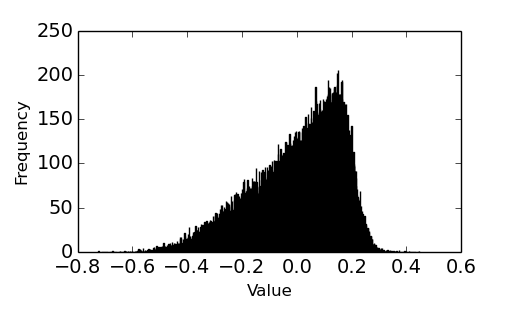}}
\hfil
\subfigure
{\includegraphics[width=1.6in]{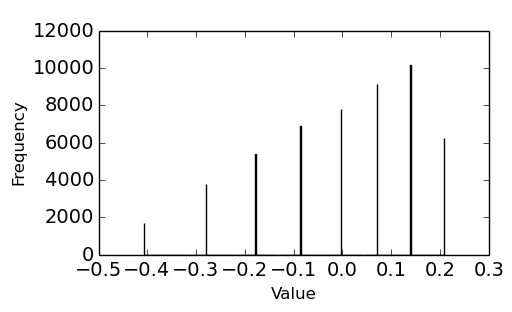}}
\hfil
\squeezeup
\\
\subfigure
{\includegraphics[width=1.6in]{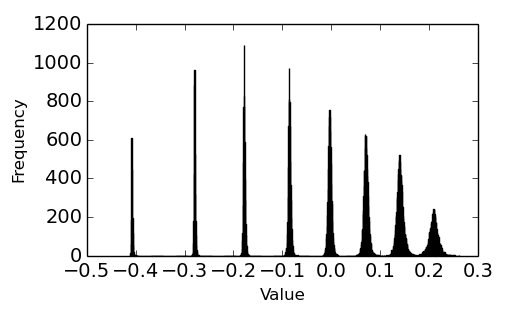}}
\hfil
\subfigure
{\includegraphics[width=1.6in]{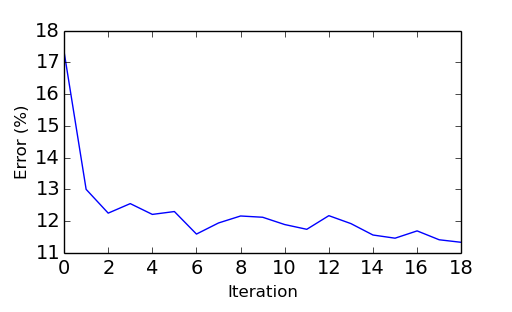}}
\squeezeup
\squeezeupless
\caption{The effect of the weight clustering on DNN weights distribution during retraining.} 
\squeezeup
\squeezeupless
\label{fig:retrain} 
\end{figure}

\noindent{\bf Weight Retraining:} Consider the distribution of the parameters within a layer shown in Figure~\ref{fig:retrain}a. Weight clustering essentially finds the best matches that can represent this distribution and replaces all parameters with their closest centroids (Figure~\ref{fig:retrain}b).
Weight clustering is often accompanied by some degree of additive error, $\Delta e=e_{clustered}-e_{baseline}$. To compensate for this error, our algorithm retrains the neural network for a pre-specified number of epochs. After retraining, the parameters have a clustered distribution as illustrated in Figure~\ref{fig:retrain}c. Therefore, a retrained weight matrix is more robust against the clustering error. The classification error decreases in subsequent clustering/retraining iterations as shown in Figure~\ref{fig:retrain}d.

\noindent{\textbf{Error Estimation}:} After the weight clustering, the error estimation module forms a software version of the reinterpreted DNN and estimates the classification error. This module replaces the original weights and neuron outputs with their closest codebook values.
The classification error $e_{clustered}$ is estimated by cross-validating the clustered DNN over a portion of the original data.
If the error rate does not satisfy the tolerance $\Delta e<\epsilon$, the model will be retrained and clustered.
This procedure is repeated for a defined number of iterations. Note that all pre-processing operations in the DNN Composer module are performed offline and their overhead will be amortized among all future executions of \Design accelerator.
In our evaluation, we empirically set the maximum number of iterations to 5 while $\epsilon$ is given by 0, to get the best model within reasonable analysis time.
We discuss the running time overhead of the whole procedure in Section~\ref{sec:expset}.

\begin{figure*}[t!]
\squeezeup
\squeezeup
\squeezeupless
  \centering
   \epsfig{file=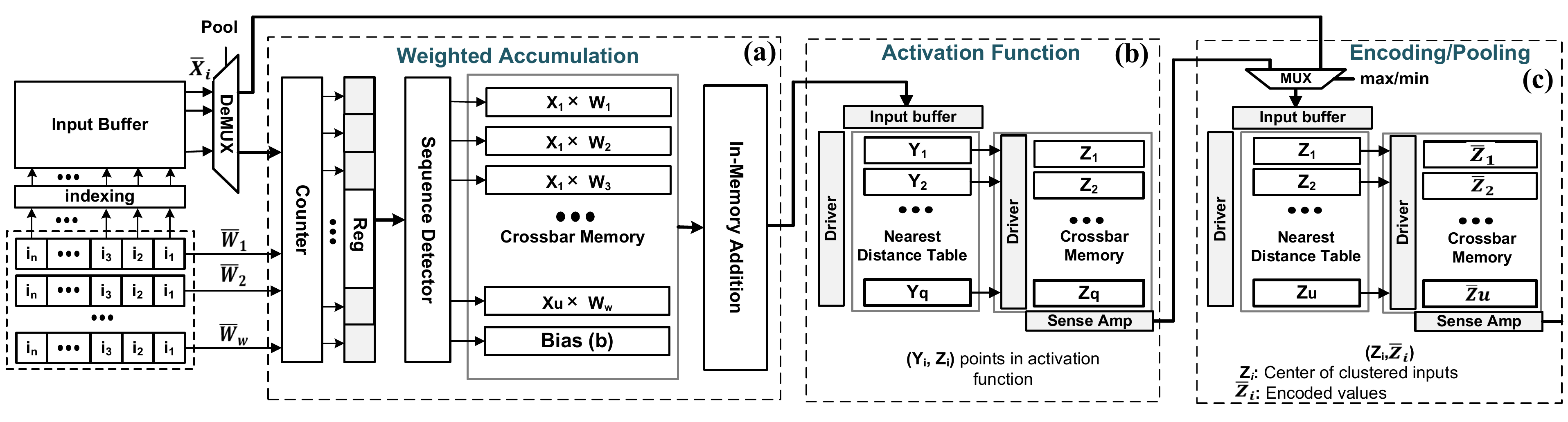, width=0.88\textwidth}
\squeezeup
\squeezeup
  \caption{An \RNA block for accelerating reinterpreted neurons.}
\squeezeup
\squeezeupless
  \label{fig:rnahardware}
\end{figure*}

\subsection{\RNA Configuration}
After retraining the networks sufficiently, we configure the reinterpreted model into the accelerator.
To write the neurons of either the fully-connected or convolution layers, an adjustable parameter is utilized to select the level of the codebook tree, i.e., the number of encoding bits.
Based on the encoding bits, we store pairwise multiplication results extracted from all possible pairs of codebook values into a crossbar memory.
The lookup tables for the quantized activation function and the encoding table are stored in two AM blocks.
As explained in Section~\ref{sec:pre}, the \textit{virtual} layer responsible for encoding the raw inputs is also stored into a AM block .
For the neurons of the pooling layer, we allocate a set of \RNA blocks.
In the next section, we explain how the \RNA memory blocks are designed to perform the computation tasks of each neuron in different types of layers.

\section{\RNA accelerator}
Figure~\ref{fig:rnahardware} illustrates the structure of an \RNA block which performs the computation tasks of a single neuron in the reinterpreted model.
An \RNA block consists of three major memristor memory blocks, (a) weighted accumulation, (b) activation function, and (c) encoding/ pooling blocks, each corresponding to one of the fundamental operations discussed in Section~\ref{sec:pre}.
The weighted accumulation sub-block is a crossbar memory capable of processing addition in-memory.
The other two sub-blocks are designed using AM structures that implement a lookup table like functionality and have the capability of searching for the most similar value in the memory.

\subsection{\RNA Weighted Accumulation}\label{sec:in_mem-add}
Since all the weights and inputs are passed to the \RNA block as encoded values, we can directly fetch the multiplication results from the crossbar memory as discussed in Section~\ref{sec:pre}.
Although our design significantly reduces the cost of multiplication, serially accumulating the values in the neuron can be a bottleneck. Weight and input clustering significantly reduces the number of possible results of multiplications. For instance, in a neuron with 1024 incoming branches, there are $w\times u$ different pre-computed values, where $w$ and $u$ are the number of codebook values for weights and inputs. Our design replaces each incoming edge of the neuron with one of the pre-computed multiplication values. As $w\times u$ is usually smaller than the number of incoming edges to the neuron, we do not need to really accumulate 1024 numbers together. Instead, using counter blocks, we record the number of times that each pre-stored value repeats. Finally, the pre-stored values are added together based on the number of times that each value occurs. This improves the performance and energy efficiency of accumulation. 

\subsubsection{Parallel Counting: }
The system introduced above can be easily implemented by having a FIFO at the input of each layer and having an increment by 1 counter corresponding to each pre-stored value. Each output of this buffer increments the corresponding counter by 1.  This procedure is highly serialized and may bottleneck the entire process. Hence, it would be beneficial to take in multiple inputs at a time and increment counters in parallel. The problem arises when two or more of these inputs correspond to the same pre-stored value. In this case, the counter would increment by just 1, resulting in erroneous results. We address this issue by exploiting the fact that each input-weight combination corresponds to a unique pre-stored value. We implement hardware such that only one input-weight pair is selected per weight at a time.

Our design assigns $w$ buffers for $w$ distinct weights. These buffers store the input indexes which use the same weight. For example, buffer corresponding to $W_0$ weight stores the indexes of all inputs to the neuron which use $W_0$ weight. The buffer size is determined by the size of the largest layer in the neural network, as this number determines the maximum incoming edges to a neuron. Our design picks one index from each weight buffer in one cycle and increments the corresponding counter. Since the input-weight combinations selected in one cycle have different weights, no two of these combinations increment the same counter.

The output of this procedure is the values of the counter which show the number of times each pre-stored value is accessed. Now, instead of repeatedly adding the numbers together, our design first shifts each pre-stored value depending upon the number of times it repeats. For instance, if the first pre-stored value repeats 4 times, our design shifts that value by two bits. The values with counters equal to 8 and 16 shift by three and four bits respectively. If the counter value is not a power of two, our design breaks the number into multiple powers of two. For example, when the counter is 9, our design breaks it to 8+1; thus the value is shifted by three bits and then added to itself. To further improve the efficiency of the process, our design tracks the longest sequence of 1s in the value of the counter and changes it to a power of 2 followed by subtraction of 1. For example, when the counter is 15 (b:1111), our design changes it to 16-1.

\subsubsection{In-Memory Addition: }~\label{sec:magic}
We break down the addition operation into a series of \verb|NOR| operations, 
where each \verb|NOR| operation in the crossbar memory is executed with a latency of 1 cycle~\cite{talati2016logic}.
Previous work has demonstrated ways, both in literature~\cite{kvatinsky2014magic5, Kvatinsky2014imply} and fabricated chips~\cite{jang2018memristive}, to implement logic using memristor switching. 
The output device switches between two resistive states, $R_{ON}$ (low resistive state, `1') and $R_{OFF}$ (high resistive state, `0'), whenever the voltage across the device exceeds a threshold~\cite{kvatinsky2015vteam2}. 
This property can be exploited to implement \texttt{NOR} gate in the digital memory by applying a fixed voltage across the memristor devices~\cite{kvatinsky2014magic5}.
To accelerate addition, our design supports addition operation in a tree structure~\cite{imani2017ultra}. As in-memory computation is slow in propagating delay, our design uses the idea of carry-save-adder to add multiple numbers together in a tree structure. This in-memory implementation can add multiple numbers in parallel while delaying the propagation to the final stage in the tree. 
For $w\times u$ inputs in a crossbar memory, our design can handle addition in $log_{3/2}(w\times u)$ stages. Each stage takes 13 cycles to complete the addition operation. Finally, the last stage requires $13N$ cycles to perform addition while propagating carry ($N$ is the size of numbers to be added).

\subsection{\RNA AM-Based Computation}
\label{sec:rnaencode}
\subsubsection{Activation Function, Encoding / Pooling:}
\label{sec:actsenpool}
The two sub blocks which implement the activation function and encoding/pooling are designed as AM blocks, i.e., lookup tables.
As shown in Figure~\ref{fig:rnahardware}b and c, an AM block has two memories, a \textit{nearest distance table} designed by a CAM structure, and a \textit{crossbar memory} which stores data associated with each row of the nearest distance table.
Since the activation function and encoding are approximately modeled by the DNN composer and stored in the AM blocks, they can be computed by activating the corresponding AM block.
In other words, the AM block for the activation function first activates its nearest distance CAM.
Then, this CAM finds the row with the data most similar to the value computed by the weighted accumulation.
The crossbar memory stores the result of the activation function which is sent to the next AM block for encoding.
Similarly, the encoding AM block produces the encoded value.

The neurons of pooling layers are implemented by reusing the last AM block which was used for the encoding task.
Since the pooling layer does not have the computation functionality, it bypasses the encoded input data, $\overline{X}_i$, to the last AM block which is then written in its CAM block.
Then we find the largest (smallest) value in the AM block if the pooling layer implements max (min) pooling.
Note that our design can also support average pooling using the weighted accumulation block. As explained in Section~\ref{sec:magic}, the crossbar memory can perform in-memory addition without the need for external circuits. The division required in average pooling is implemented by normalizing the weights in the offline stage.
In the following subsection, we explain how we design the nearest distance table using a CAM, called \ReCA. 

\subsubsection{Nearest Distance CAM:}
\label{sec:NNCAM}
A conventional CAM design finds the exact same data as given input data.
As discussed in Section~\ref{sec:Related}, there are some NVM-based designs that allow the search for a ``similar'' data.
To quantify this ``similarity'', there exist different metrics such as hamming distance and absolute distance. 
The Hamming distance (HD) is one of the simplest distance metrics which can be implemented in the memory in a relatively easy way.
However, this metric ignores the impact of the bit indices on the computation.
For example, $11111$ has the same HD to $11110$ and $01111$, while the absolute distances in numeric values are significantly different.
In this work, we first show how to design a CAM with the capability of searching for the nearest HD value.
Then, we present how to make a modification on lookup circuits to enable a precise search operation in \ReCA which identifies the value with the smallest absolute distance for real numbers.

\begin{figure}
\squeezeup
  \centering
\centerline{
{\includegraphics[width=1\linewidth]{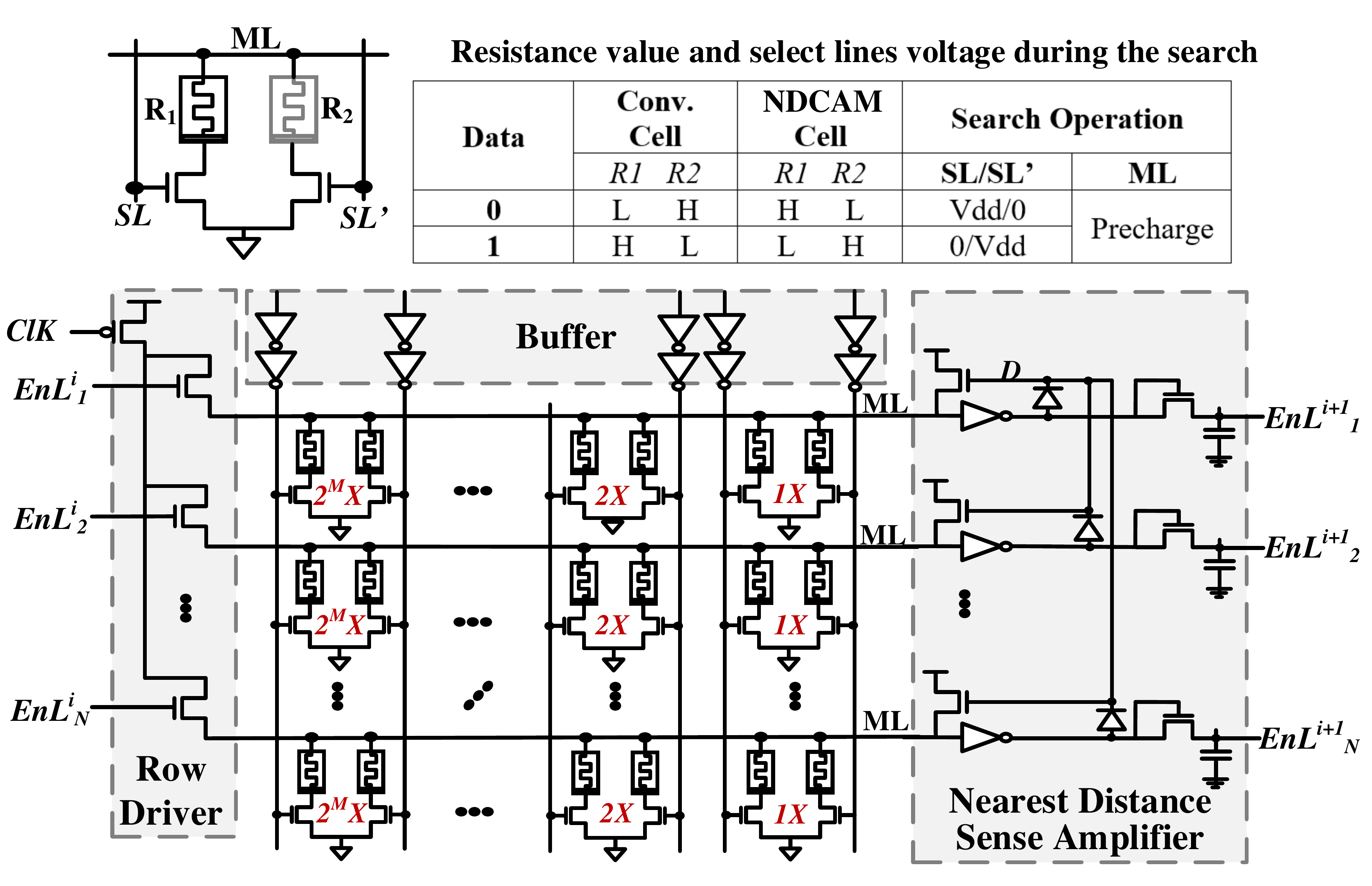}}
}
\squeezeup
\squeezeupless
\caption{\ReCA supporting nearest distance search.}
   \squeezeup
\squeezeupless
\squeezeupless
  \label{fig:CAMCell}
\end{figure}

\noindent{\bf \ReCA Search Functionality:}
Figure~\ref{fig:CAMCell} shows the structure of our \ReCA design.
Before the search operation, the input data is stored in the buffer, and the buffer strengthens the input signals to ensure every row can receive the input signals at the same time.
A typical way to differentiate the HDs of stored values to the input signal is to exploit a timing characteristic of the discharging current for each row~\cite{guo2015resistive, rahimi2015approximate}.
%
In this approach, for the search operation, match lines (ML) of all rows are precharged to $Vdd$.
Then, if the bit stored in each cell is different from the input signal, the corresponding \textit{ML} starts discharging. 
For a large number of mismatched bits, the rows discharge \textit{ML} voltage with higher current and at a faster rate compared to other rows with smaller mismatched bits.
Thus, a sense amplifier can detect the CAM row which lastly discharges, i.e., the value with the nearest HD, by keeping track of \textit{ML} voltages in all rows.
However, this approach makes the sense amplifier complicated due to the additional circuity such as counters.
In addition, it needs to wait for a long time to determine the row lastly discharged.

To address these design issues, the CAM cells in proposed \ReCA work inversely compared to the typical CAM.
The table shown in Figure~\ref{fig:CAMCell} presents the functionality of \ReCA cells storing inverse resistance values in the match and mismatch cases.
In contrast to the conventional cells, \ReCA cell discharges the \textit{ML} in case of matching, while a mismatch \textit{ML} stays charged.
Therefore, a row which has more matched bits creates a faster discharging current than other rows.
The inverse mode simplifies the sense amplifier design to detect the nearest HD row, since we only need to find the row which discharges the \textit{ML} fastest. 
On the top of the inverse scheme, we modify the CAM design to support the precise search operation which identifies the row with the smallest absolute distance.
To this end, each CAM for different bit indices is designed using different access transistor sizes.
Based on the binary weight of an unsigned integer value, each cell in a $i^{th}$ position has access transistors which are $2\times$ larger than the cell in the $i-1^{th}$ adjacent bit.
This results in $2\times$ higher \textit{ML} discharging current in each match cell than its adjacent least significant bit (LSB).

In fact, the number of block bits, and the size of transistors and capacitors affects the timing characteristic.
Thus, we identified viable configurations so that they can guarantee the correct functionality even for the worst case. 
In our HSPICE evaluation of 5000 Monte Carlo simulations considering 10\% of process variation,
the discharging speed is sufficiently distinguishable when an ML has 8 subsequent bits.
Thus, we divide 32 bits into 4 pipeline stages and find the closest row by performing sequential search starting from most significant bits. 
A CAM block only includes 8 bits, and thus the access transistors can be a reasonable size even for the MSB of a stage.
To support floating point data, we put the exponent and fraction parts in different stages.  
NDCAM performs any activation/pooling functions in a single-cycle using the search operation. For example, to implement $4\times 4$ MAX pooling, NDCAM requires $24\mu m^2$ area, $0.5ns$ search latency, and $920fJ$ energy. Running the same function on CMOS requires $374\mu m^2$ area, $1.2ns$ latency, and $378fJ$ energy.

\subsection{\Design Data Transfer}
\begin{figure}
\squeezeup
\squeezeup
\squeezeupless
  \centering
\centerline{
{\includegraphics[width=1\linewidth]{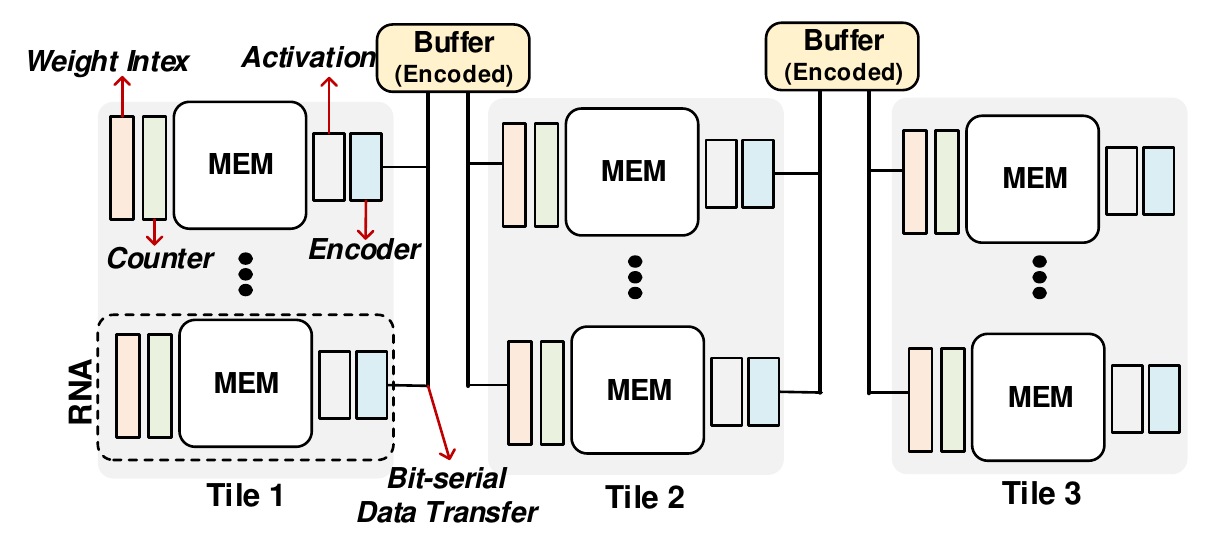}}
}
\squeezeup
\squeezeupless
\caption{\Design architecture overview.}
   \squeezeup
\squeezeupless
  \label{fig:architecture}
\end{figure}

Figure~\ref{fig:architecture} shows the overview of the \Design architecture modeling multiple layers of neural networks. \Design consists of several blocks working in parallel to model the computation of different DNN layers. 
In \Design, each block consists of 1k RNA blocks are working in parallel. The outputs of these RNAs are written in parallel into a single buffer. 
This buffer values are the encoded outputs of a DNN layer which are used as input data for the neuron of the next layer. All RNAs access to the buffer values in parallel. 
The data transfer from the neurons to buffer happens in a bit serial way. Since the values are encoded, this data transfer can perform significantly faster than the original 32-bits numbers. 
\Design works in a pipeline, meaning that when a block is writing values into a buffer, the next block (next layer) accessing the previous values stored in the buffer. This pipeline structure maximizes  \Design throughput. 


\section{Experimental Results} \label{sec:results}
\subsection{Experimental Setup}
\label{sec:expset}
The proposed \Design framework has been implemented with the two co-designed modules, DNN composer for software and accelerator for hardware.
We designed the DNN composer, which retrains DNN models for the accelerator configuration, in C++ while exploiting two backends, Scikit-learn library~\cite{pedregosa2011scikit} for clustering and Tensorflow~\cite{chollet2015,abadi2016tensorflow} for the model training and verification. 
For the accelerator design, we exploit HSPICE design tool for circuit-level simulations and calculate energy consumption and performance of all the \Design memory blocks. The energy consumption and performance is also cross-validated using NVSim~\cite{dong2014nvsim}.
The \Design controller has been designed using System Verilog and synthesized using \textit{Synopsys Design compiler} in 45nm TSMC technology.

\begin{table}
\caption{\Design Parameters}
\resizebox{1\columnwidth}{!}{
\begin{tabular}{c|ccc||c|ccc}
\hline
 \multicolumn{4}{c||}{\textbf{1-RNA Block}}                           & \multicolumn{4}{c}{\textbf{1-Tile}}                                                                           \\ \toprule
\textit{Blocks}                 & \textit{Size}   & \textit{Area} & \textit{Power} & \textit{Blocks}               & \textit{Size}     & \textit{Area} & \textit{Power (w)} \\ \midrule
\textit{\textbf{Crossbar}}      & 1K*1K           & 3136$\mu m^2$            & 3.7mW               & \textit{\textbf{RNAs}}        & 1k                & 3.84 $mm^2$                              & 4.8W               \\ 
\textit{\textbf{Counter}}       & 1k*12-bits      & 538.6$\mu m^2$           & 0.7mW               & \textit{\textbf{Buffer}}      & 1K-reg          & 37.6$\mu m^2$                                & 2.8mW              \\  \cline{5-8}
\textit{\textbf{Activation}}    & 64-rows         & 83.2$\mu m^2$            & 0.2mW               & \multicolumn{2}{c}{\textit{\textbf{Total Tile}}} & \textbf{3.88}$mm^2$                       & \textbf{4.8W}      \\ \cline{5-8} \cline{5-8}
\textit{\textbf{Encoder}}       & 64-rows         & 83.2$\mu m^2$            & 0.2mW               & \multicolumn{4}{c}{\textbf{Total Chip}}                                                                             \\ \hline 
\multicolumn{2}{c}{\textit{\textbf{Total RNA}}} & 3841$\mu m^2$            & 4.8mW               & \multicolumn{2}{c}{\textit{\textbf{32-Tiles}}}   & \textbf{124.1$mm^2$}                      & \textbf{310.4W}    \\ \hline
\end{tabular}
}\label{tab:config}
\squeezeup
\end{table}

One major advantage of \Design is that it can work with any bipolar resistive technologies which are the most commonly used in existing NVMs. Here, we adopt a memristor device with a large OFF/ON resistance~\cite{Kvatinsky2015vteam} for the memory devices. The robustness of all proposed circuits has been verified by considering 10\% process variations on the size and threshold voltage of transistors using 5000 Monte Carlo simulations. 
We compare the proposed \Design accelerator with GPU-based DNN implementations, running on NVIDIA GTX 1080 GPU. All DNN applications are realized using Tensorflow~\cite{abadi2016tensorflow} and the GPU time and power are measured using the \verb|nvidia-smi| tool. 

Table~\ref{tab:config} shows the details of \Design parameters consisting of 32 Tiles. Each tile consists of 1k RNA blocks and a single buffer storing intermediate input/output results. Each RNA has crossbar memory, counter, activation, and encoder blocks. \Design totally consumes $155.3W$ maximum power and takes $124.1mm^2$ area.


\begin{table}[t!]
\squeezeup
\squeezeup
\squeezeupless
\centering
\caption{DNN models and baseline error rates (Input - $IN$, Fully connected - $FC$, Convolution- $CV$, and Pooling layers - $PL$.)}
\label{tab:applications}
\resizebox{\columnwidth}{!}{
\begin{tabular}{c|l|l}
\hline
\textbf{Dataset   }                      & \multicolumn{1}{c|}{\textbf{Network Topology}}                                                                                                                                            & \multicolumn{1}{c}{\textbf{Error}} \\ \hline
\textbf{MNIST }                          & $IN:784,\ FC:512,\ FC:512,\ FC:10$                                                                                                                                                   &  1.5\%                                  \\ \hline
\textbf{ISOLET  }             & $IN:617,\ FC:512,\ FC:512,\ FC:26$                                                                                                                                                  & 3.6\%                                \\ \hline
\textbf{HAR}            & $IN:561,\ FC:512,\ FC:512,\ FC:19$                                                                                                                                                   &  1.7\%                                   \\ \hline
\multicolumn{1}{c|}{\textbf{CIFAR-10}}  & \multirow{2}{*}{\begin{tabular}[c]{@{}l@{}} $IN:32\times 32\times 3,CV:32\times 3\times 3,PL:2\times 2,$\\$CV:64\times 3\times 3,CV:64\times 3\times 3,FC:512,\ FC:10\ (100)$ \end{tabular}} &   14.4\%                                  \\ \cline{1-1} \cline{3-3} 
\multicolumn{1}{c|}{\textbf{CIFAR-100}} &                                                                                                                                                                                  &  42.3\%                                   \\ \hline
\multirow{3}{*}{\textbf{ImageNet} } & 
\begin{tabular}[c]{@{}l@{}} AlexNet~\cite{krizhevsky2012imagenet} \end{tabular} &  43.0\%                                  \\ 
& \begin{tabular}[c]{@{}l@{}} VGG-16~\cite{simonyan2014very} \end{tabular} &  28.5\%                                  \\ 
& \begin{tabular}[c]{@{}l@{}} GoogleNet~\cite{szegedy2015going} \end{tabular} &  15.6\%                                  \\ 
\hline
\end{tabular}
}
\squeezeup
\end{table}

\begin{figure*} [t]
\squeezeup
\squeezeup
\squeezeupless
\begin{minipage}{1.0\textwidth}
\end{minipage}
\centerline{
{\includegraphics[width=7.22in]{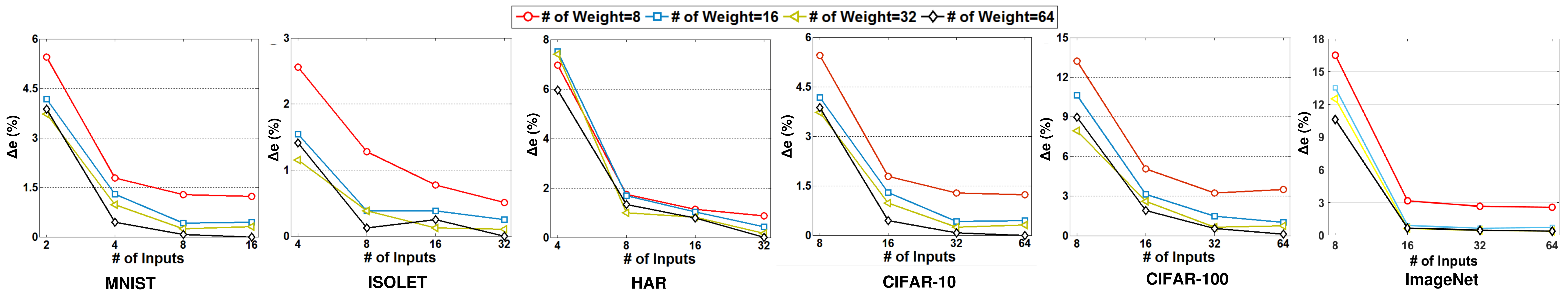}}
}
\squeezeup
\caption{Accuracy loss of the model reinterpretation for different sets of input and weight sizes.}
\squeezeup
\squeezeupless
\label{fig:Accuracy2}
\end{figure*}

\subsection{Benchmarks and DNN Models}
We evaluate the efficiency of the proposed \Design over six popular neural network applications:
Handwriting classification (MNIST)~\cite{MNIST}, Voice Recognition (ISOLET)~\cite{Isolet}, Activity Recognition (HAR)~\cite{HAR}, Object Recognition (CIFAR)~\cite{cifar}, and Image Classification (ImageNet)~\cite{imgnet}
The Table~\ref{tab:applications} also presents the DNN topologies and baseline error rates for the original models before reinterpretation. As for well-known applications such as CIFAR, we have used the architecture suggested by the Keras library. The pretrained baselines for ImageNet, including AlexNet~\cite{krizhevsky2012imagenet}, VGG-16~\cite{simonyan2014very}, and GoogleNet~\cite{szegedy2015going} architectures, are taken from the Keras library as well. For other applications, we chose the network architecture that achieves fairly high baseline accuracy (e.g., standard 98.4\% for MNIST without convolutions). 
The error rate is defined by the ratio of the number of misclassified data to the total number of a testing dataset.
Each DNN model is trained using stochastic gradient descent with momentum~\cite{sutskever2013importance}.
In order to avoid overfitting, Dropout~\cite{srivastava2014dropout} is applied to fully-connected (FC) layers with a drop rate of 0.5.
In all the DNN topologies, the activation functions are set to ``Rectified Linear Unit'' (ReLU) for hidden layers, and a ``Softmax'' function is applied to the output layer. 
.

\subsection{Accuracy of Reinterpreted DNN Models}
\label{sec:accuracy}
As we discussed previously in Section~\ref{sec:qual}, the accuracy of the model increases for a higher number of retraining epochs. Although the runtime overhead of model reinterpretation amortizes across all future executions of \Design, one might question the relative cost of reinterpretation compared to the initial training phase. As such, we deliberately limit the number of retraining epochs to 1 for Imagenet and 5 for the other datasets to ensure that the reinterpretation overhead is negligible compared to the actual training.

\begin{figure*} [t!]
\begin{minipage}{1.0\textwidth}
\centerline{
{\includegraphics[width=0.85\textwidth]{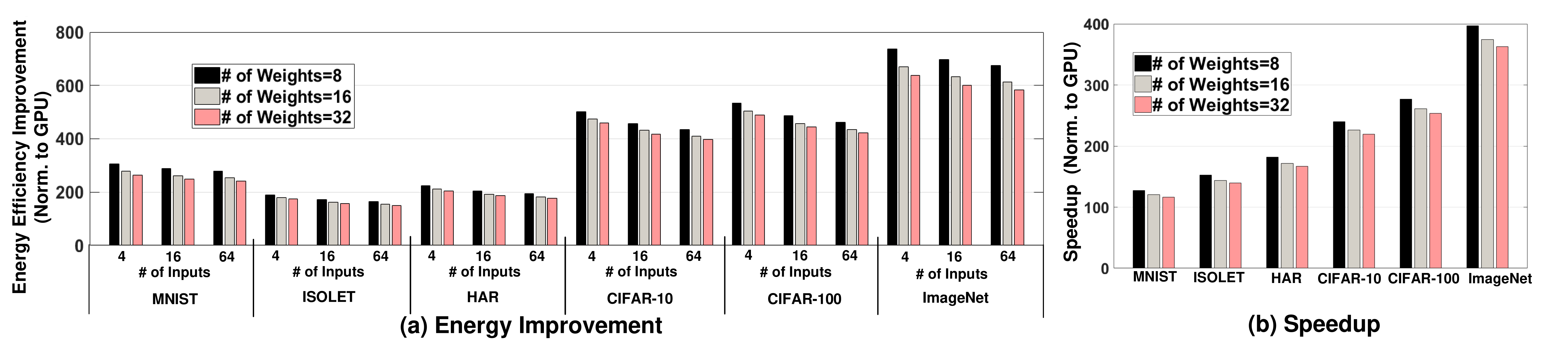}} }
\end{minipage}
\squeezeupless
\squeezeup
\caption{Energy efficiency and performance improvement of 6 models normalized to GPU-based executions.}
\squeezeup
\squeezeupless
\squeezeupless
\label{fig:compare}
\end{figure*}

As for the hardware accelerator, the accuracy of the reinterpreted model is affected by three major configurable factors: (i)~the number of quantized values for an activation function ($q$), (ii)~the number of clustered weights ($w$), and (iii)~the number of clustered inputs ($u$).
They also decide memory sizes and consequent power/performance efficiency of the accelerator.
Since we use the same lookup table for the activation functions over all RNNs, we first show accuracy changes for different $q$ to select a proper configuration. To evaluate the accuracy of our reinterpreted models, we exploit the $\Delta e$ accuracy loss metric defined in Section~\ref{sec:qual}, i.e., how much the error is changed over the baseline error rate.
Our evaluation shows that for all benchmarks, using lookup table with 64 rows to modify activation function (Sigmoid) results in the same accuracy level to the baseline models which exactly compute the activation function results. Note that for ReLU function, it is simpler and more efficient to design it using a single CMOS comparator.

Figure~\ref{fig:Accuracy2} shows the impact of $w$ and $u$ (i.e., the number of the representative weights and inputs obtained from the clustering respectively) on the inference accuracy of the six benchmarks. For each dataset we have shown the result for a single network. For ImageNet, the results are shown for VGG-16 network. 
We changed the numbers by selecting a tree level for each codebook.
The results show that exploiting more clusters provides better accuracy in general.
When clustering with 16 and 64 for the weights and inputs, the reinterpreted models achieve the same accuracy level, i.e., $\Delta e \approx 0\%$, for most applications.
We observe that different benchmarks require different cluster numbers to provide acceptable quality.
For example, the DNN model for MNIST is performed with $\Delta e = 0$ when $w=64$ and $u=16$.
In contrast, the ImageNet, which are known as a more complex classification task, requires 64 clustered weights and 64 clustered inputs to provide similar quality to the baseline. Our evaluation shows that for AlexNet, VGG-16, and GoogleNet, \Design provides less than 0.1\%, 0.3\%, and 0.5\% quality loss using 64 clustered inputs/weights.
In the following subsection, we show how the number of clustered values affect \Design efficiency by determining the size of the crossbar array storing pre-computed multiplication results and the size of encoding AM block.

\begin{figure*} [t!]
\squeezeup
\squeezeup
\squeezeupless
\begin{minipage}{1.0\textwidth}
\centerline{
{\includegraphics[width=0.85\textwidth]{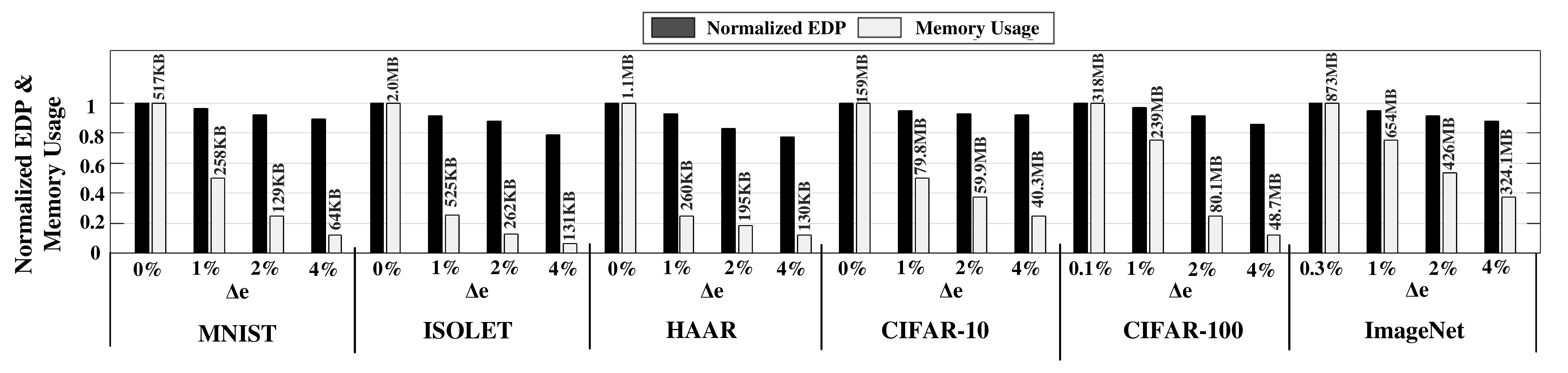}} }
\end{minipage}
\squeezeupless
\squeezeup
\caption{Normalized energy-delay product and memory usage of \Design for different accuracy levels.}
\squeezeup
\squeezeupless
\label{fig:approx}
\end{figure*}

\subsection{Accuracy-Efficiency Trade-off}
\label{sec:expeffieicncy}
Figure~\ref{fig:compare} shows energy improvement and performance speedup of the six applications running on the proposed \Design and the GPU implementation. 
We consider the efficiency for 9 combinations of different cluster sizes, where either input or weight are encoded (clustered) with 4, 16 and 64 values.
The results show that the \Design accelerator improves the energy and performance efficiency significantly compared to the GPU-based implementation.
Comparing with GPU, the speedup stems from the fact that \Design offers much higher parallelism by (i)completely parallelizing each neuron computation with RNAs,(ii) ensuring each RNA to store the weights of the corresponding neuron. \Design can perform 10 million operations in parallel, while for GPU it is in order of thousands. 

In \Design, the energy and performance efficiency is mainly related to two factors: i) the size of the multiplication crossbar memory affected by both the $w$ and $u$, and ii) the size of the encoding AM block affected by $u$.
Since $u$ affects the two different memory blocks, the number of encoded inputs has a higher impact on energy consumption than the number of the encoded weights.

In addition, the number of the encoded weights has negligible impacts on performance as we can extract a multiplication result by directly referring a row of the crossbar memory. We report the speedup for different $u$ values in Figure~\ref{fig:compare}b.
The efficiency improvement depends on the combination, that is, using smaller encoded input and weight sets results in more energy-efficient and faster computation.
For example, we achieve $253.2\times$ energy efficiency improvement and $422.5\times$ speedup for $w=4$ and $u=4$, whereas $161.9\times$ and $386.25\times$ for energy and performance when $w=64$ and $u=64$.

The memory sizes also affect the model accuracy as well as the accelerator efficiency.
To evaluate the relationship, we chose four accuracy loss values, i.e., $\Delta e$, from minimum to 4\%, and selected a combination whose energy-delay product (EDP) is minimal for each accuracy loss over all applications.
Figure~\ref{fig:approx} summarizes the EDP normalized to the case with minimum $\Delta e$ along with its memory usage for different accuracy levels. 
The results show that by allowing small accuracy loss, we could achieve better EDP efficiency.
For example, for the $\Delta e= 2\%$ and $4\%$ cases, the \Design acceleration can save EDP by 11\% and 15\% respectively, as compared to minimum $\Delta e$ case.
This also allows to use less memory of the accelerator, e.g., 77\% and 87\% for  $\Delta e= 2\%$ and $4\%$ cases.

Note that our reinterpreted model effectively enables PIM-based computing with a relatively small amount of memory usage while completely removing the need for ADC and DAC on the PIM-based DNN acceleration.
The largest memory usage is observed for ImageNet and CIFAR-100, by 837MB and 318MB with minimal loss of the inference quality of 0.3\% (VGG-16) and 0.1\% respectively.
In addition, since each application requires different memory sizes for the best configuration, a system designer may configure the accelerator depending on the running application by choosing the level of the codebook which decides the number of encoded weights and inputs.

\subsubsection{Energy/Performance Breakdown:}
\begin{figure}[t!]
  \centering
   \epsfig{file=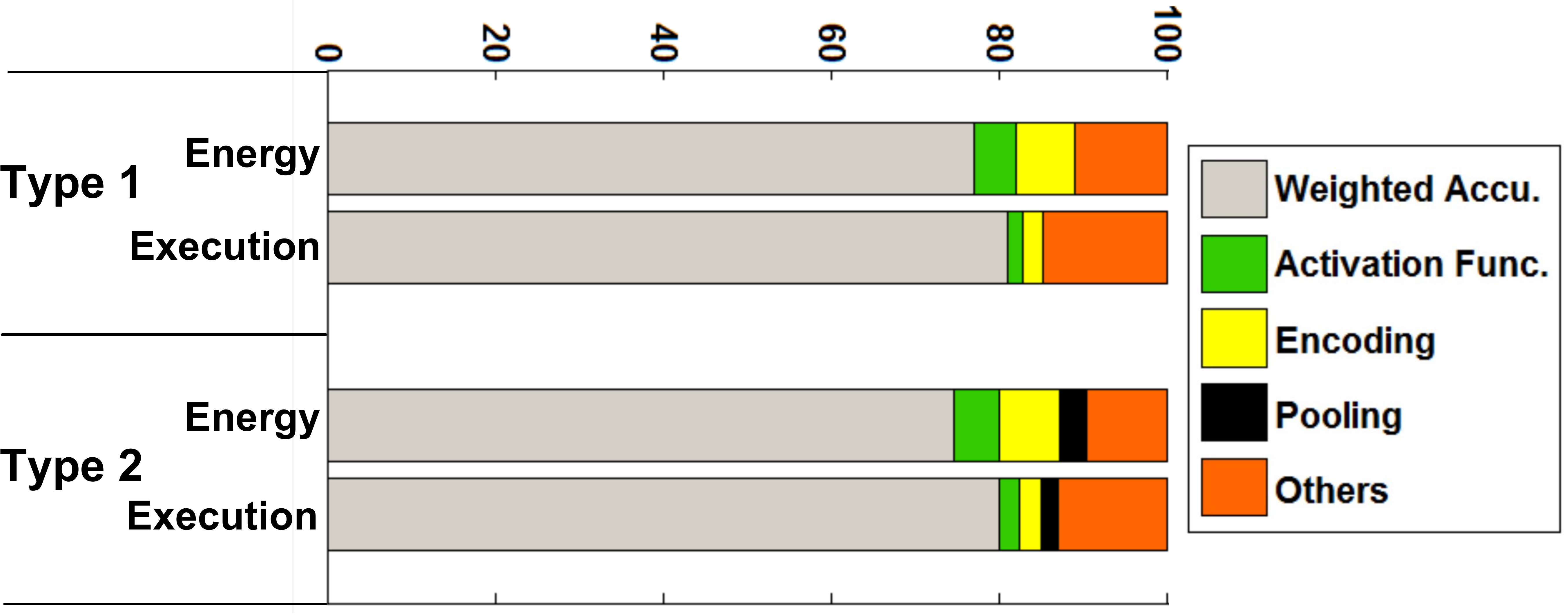, width=0.8\columnwidth}
\squeezeup
\vspace{-1mm}
\caption{Breakdown of energy and execution time.}
  \squeezeup
  \squeezeupless
  \squeezeupless
  \label{fig:Breakdown}
\end{figure}

To further analyze how the proposed accelerator consumes energy and performance, we classified the energy consumption and execution time for the three major memory blocks, i.e., weight accumulation, activation function, encoding/ pooling, and other hardware blocks, when $w=u=64$.
According to the model topology, we defined two groups for the six applications, (i) \textsf{Type 1}, whose models consist of fully connected layers (MNIST, ISOLET, and HAR), and (ii) \textsf{Type 2}, whose models consist of fully connected, pooling, and convolution layers (CIFAR-10, CIFAR-100 and ImageNet).
Figure~\ref{fig:Breakdown} shows the breakdown for the two application groups.
The results show that the memory block for the weighted accumulation consumes a dominant portion of the energy and execution time for the two types, 77.1\% and 81.4\%, respectively, as the multiplication and addition are the most frequent operations in the neural networks.
In contrast, the two memory blocks for the activation function and encoding takes less portion since the AM blocks that support nearest distance searches can efficiently identify the desired data.
The pooling neurons are used only in \textsf{Type 2} models to process the outputs of convolution layers.
This block consumes 3.2\% of the energy and 1.9\% of the execution time.
The other hardware blocks, including a broadcast buffer and a memory controller, MUXs, and address decoders, take about 11.2\% and 14.8\% for the energy and execution time, respectively, while the majority is consumed by the broadcast buffer (69\% and 75\% within the sub-portion).

\squeezeup
\subsubsection{\Design Area Analysis:}
\begin{figure}[t!]
\squeezeup
\squeezeup
\squeezeupless
  \centering
  \epsfig{file=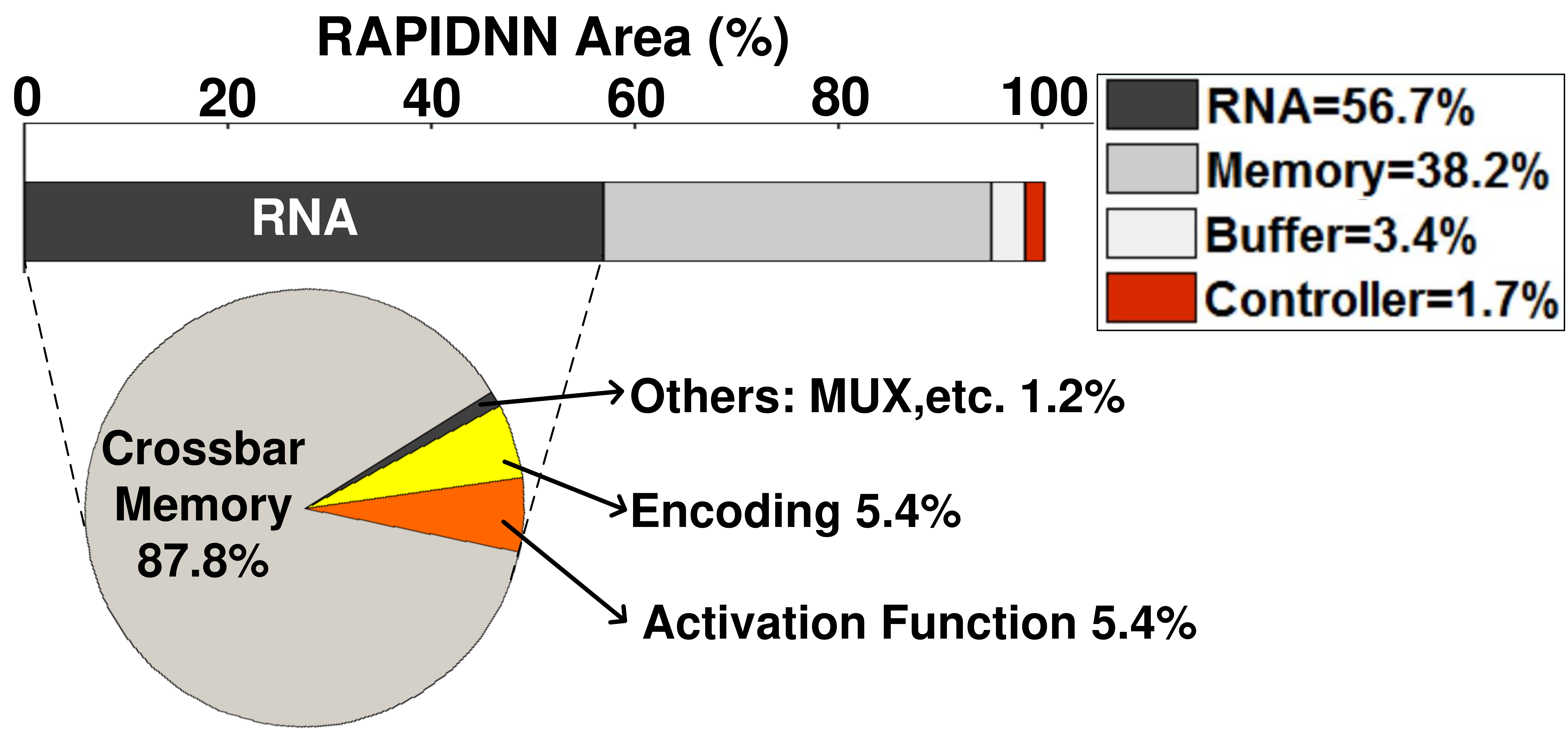, width=0.8\columnwidth}
    \squeezeupless
     \squeezeup
  \caption{\Design area breakdown.}
     \squeezeupless
     \squeezeup
     \squeezeupless
  \label{fig:area}
\end{figure}

\Design provides a significant improvement in area efficiency as compared to prior accelerators because: (i) \Design does not need to store all weights but just the multiplication results of clustered inputs/weights in small memory. (ii) \Design works in a digital domain using a binary representation and does not require ADC/DAC blocks which take the majority of the area in other in-memory accelerators such as ISAAC. Our evaluations show that \Design with $w=16$ and $u=32$ consumes 34\% less area as compared to ISAAC.
We have also analyzed how different blocks utilize the area of the \Design accelerator.
Figure~\ref{fig:area} shows that the \RNA and memory blocks take 56.7\% and 38.2\% of the total area, respectively.
The rest of 5.1\% area corresponds to the buffer and controller block.
The area of an \RNA block is divided into four parts, (i) a crossbar memory for storing multiplication results, (ii) an AM block for activation function, (iii) another AM block for encoding, and (iv) other circuits, e.g., MUX.
This analysis shows that, since the area overhead to implement the lookup table functionality in \ReCA is negligible; thus the two AM blocks take a small portion, i.e., 10.8\%, over the entire area of the \RNA.

\subsubsection{\Design Scalability:}
The evaluation results of this paper (i.e., area, energy, and runtime) are reported for fully-parallel execution in each layer. In the fully-connected layers, for instance, each output neuron has its own hardware \RNA block. This approach increases the throughput at the cost of higher power, energy, and area. In a resource-constrained setting, however, such extreme parallelism might not be feasible due to physical hardware limitations. We argue that \Design can address this issue by sharing a single \RNA block across multiple output neurons. Particularly, all output neurons of a fully connected layer have lookup tables with the exact same entries; therefore, a single \RNA block can be reused to compute the output of all neurons of the same layer. In convolution layers, all neurons of a single output channel have the same lookup table. As a result, \Design offers a tradeoff between runtime and hardware implementation costs such as power, area, and energy consumption. 
\subsection{Comparison with Existing Techniques}
The idea of weight sharing was originally proposed by~\cite{han2015deep,chen2015compressing}, where the retraining phase directly trains the shared weights by gradient averaging. Our proposal is different in that it does not use gradient averaging during the retraining, which allows us to maintain accuracy with fewer iterations (e.g., 1 epoch for ImageNet). In addition, previous works do not provide dynamically reconfigurable codebooks, for which we propose the hierarchical tree structure in Section~\ref{sec:clustering}. Finally, existing compression methods only encode the weight parameters which are stationary during the training. Our proposal also addresses the dynamic encoding of activation functions during execution. Note that, without encoding the activation functions, the idea of computing with lookup tables cannot be implemented. Another significant advantage of \Design over prior PIM-based accelerators is its easy integration using reliable single-level memristor devices, e.g., Intel 3D Xpoint. \Design exploits crossbar memory capable of in-memory addition and CAM blocks, which have been already fabricated by several works from the industry/academia~\cite{jang2018memristive, li20141}.

Here, we compare the energy and performance efficiency of \Design with the state-of-the-art DNN accelerators: DaDianNao~\cite{chen2014dadiannao}, ISAAC~\cite{shafiee2016isaac}, and PipeLayer~\cite{song2017pipelayer}. 
For these accelerators, we select the best configuration reported in the papers~\cite{shafiee2016isaac,chen2014dadiannao, song2017pipelayer}. 
DaDianNao works at 600MHz, with 36MB eDRAM size (4 per tile), 16 neural functional units, and 128-bit global bus. 
ISAAC design works at 1.2GHz and uses 8-bits ADC, 1-bit DAC, 128$\times$128 array size where each memristor cell stores 2 bits. PipeLayer works with the same configuration as ISAAC, but uses a spike-based approach for the analog matrix multiplication ($\lambda=4$). 
Here, we consider \Design in two configurations: 1-chip configuration, and  8-chips that provides the similar area as ISAAC and PipeLayer accelerators.
For each application, we set the lookup table size to ensure \Design works with near-zero accuracy loss (maximum $\Delta e=0.5\%$ for ImageNet).

\begin{figure} [t!]
\squeezeup
\squeezeup
\squeezeupless
\begin{minipage}{1.0\columnwidth}
\centerline{
{\includegraphics[width=1\columnwidth]{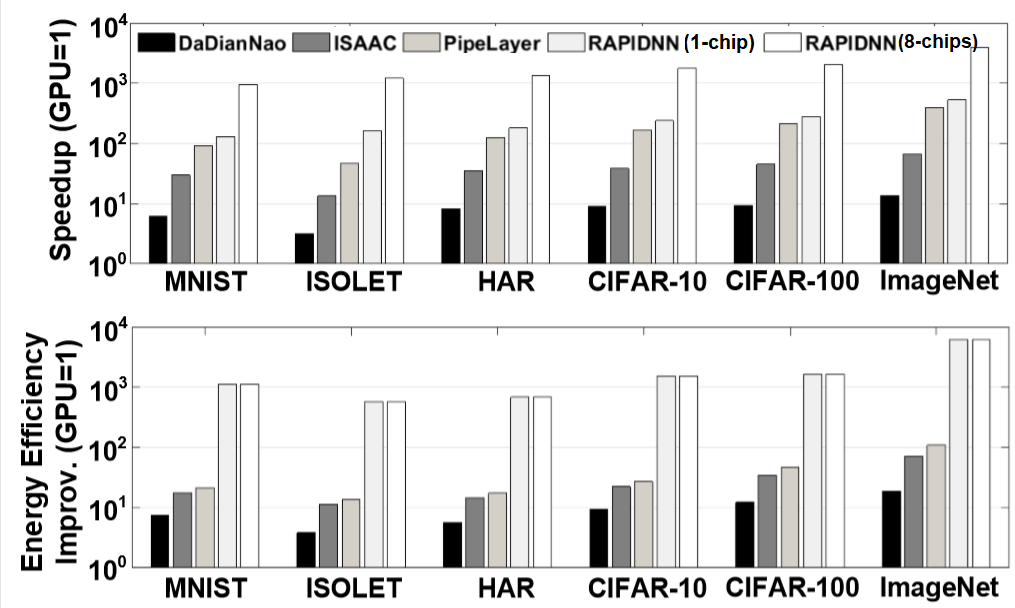}} }
\end{minipage}
\squeezeup
\caption{Comparison of \Design efficiency.} 
\squeezeupless
\squeezeup
\squeezeupless
\label{fig:existing_compare}
\end{figure}

Figure~\ref{fig:existing_compare} shows the speedup and energy efficiency improvement of different accelerators normalized to the GPU-based implementation. 
Our evaluation shows that at a similar level of accuracy, \Design using 1-chip can achieve 24.3$\times$, 5.6$\times$ and 1.5$\times$ speedup and 40.3$\times$, 13.4$\times$ and 49.6$\times$ energy efficiency improvement as compared to DaDianNao, ISAAC, and PipeLayer accelerators respectively, by hiding the data movement completely and significantly decreasing the NN computation cost.
\Design using 8-chips can further improve the computation speedup by increasing the number of \RNA blocks. Our evaluation shows that 8-chips provides 48.1$\times$, 10.9$\times$ speedup and 68.4$\times$, 49.6$\times$ energy efficiency improvement as compared to ISAAC and PipeLayer while providing a similar chip area and classification accuracy. 

In terms of computation efficiency, \Design can provide 1,904.6 $GOP/s/mm^2$ which is higher then ISAAC (479.0 $GOPS/s/mm^2$) and PipeLayer (1,485.1 $GOPS/s/mm^2$). The \Design efficiency comes from its higher density which enables more number of computations happen in the same memory area.
For example, ISAAC uses large ADC and DAC blocks which take a large portion of the memory area. In addition, Pipelayer still requires to generate spike which results in lower computation efficiency. 
\Design also can provide 839.1 $GOP/s/W$ power efficiency which is higher than both ISAAC (380.7 $GOPS/s/W$) and PipeLayer (142.9 $GOPS/s/W$). \Design removes the necessity of the costly internal data movement between the \Design blocks by using the same memory block for both storage and computing. 

%

\section{Related Work}
\label{sec:Related}
\ifx{
\subsection{Processing In-Memory}

\textbf{Non-volatile Memory:} High density, low-power consumption, and CMOS-compatibility of emerging non-volatile memories (NVMs), in particular memristor devices, make them appropriate candidates for both storage and computing purposes~\cite{guo2011resistive, yavits2015resistive, bojnordi2016memristive}. Many logic families have been proposed for computation inside the memristive crossbar. Some of these logic implementations such as stateful implication logic~\cite{borghetti2010memristive, Kvatinsky2014imply} and Memristor Aided loGIC (MAGIC) \cite{kvatinsky2014magic5} are purely realized within memory. 
The direct application of these schemes in data-intensive applications such as DNNs is highly limited due to the linear dependency of execution latency on the size of the data. While~\cite{Siemon2015} presents a very fast adder, the area overhead involved in arrayed addition grows significantly for data-intensive workloads. We use \verb|MAGIC NOR| ~\cite{kvatinsky2014magic5} to execute logic functions in memory due to its simplicity and independence of execution from data in memory. An execution voltage, $V_0$, is applied to the bitlines of the inputs (in case of \verb|NOR| in a row) or wordlines of the outputs (in case of \verb|NOR| in a column) in order to evaluate \verb|NOR|, while the bitlines of the outputs (\verb|NOR| in a row) or wordlines of the inputs (\verb|NOR| in a column) are grounded. The work in~\cite{talati2016logic, imani2017ultra} extend this idea to implement adder in a crossbar. }\fi

Modern neural network algorithms are executed on diverse types of processors such as GPU~\cite{bhuiyan2010acceleration, ciresan2011flexible}, FPGAs~\cite{gupta2015deep, sharma2016high, zhang2015optimizing, ma2017optimizing} and ASIC chips~\cite{chen2014diannao,luo2017dadiannao,ciresan2011flexible,  han2016eie, chen2017eyeriss}. 
Prior works attempt to fully utilize existing cores to accelerate neural networks.
Several prior works showed that hardware-based accelerations could further improve the efficiency of neural networks~\cite{chen2014diannao, chen2014dadiannao, aklaghi2018snapea, hegde2018ucnn, nazemi2018nullanet}.
However, the main computation still relies on CMOS-based cores, thus suffering from the data movement and lack of parallelism. 

To address data movement issue, prior works accelerate neural network by enabling analog-based PIM operations~\cite{song2017pipelayer, cheng2017time, cai2018training, cai2018long}.
Work in\cite{bojnordi2016memristive,bojnordi2017memristive} designed NVM-based Boltzmann machine capable of solving a broad class deep learning and optimization problems. 
Work in~\cite{chi2016prime,shafiee2016isaac} used ReRAM-based crossbar memory to perform matrix multiplication in memory and accordingly designed architecture to design PIM-based accelerator for CNN inference.
 Work in~\cite{feinberg2018enabling} extended the analog-based PIM to support floating point operations.
Work in~\cite{fujiki2018memory} generalized the idea of analog-PIM to accelerate general applications by offloading the PIM-compatible operations.   
However, all these approaches have potential design issues: first, their designs require to use ADC/DAC blocks, which dominate the chip area/power~\cite{shafiee2016isaac}. 
Second, they use multi-level memristor devices that are not sufficiently reliable for commercialization unlike commonly-used single-level NVMs, e.g., Intel 3D Xpoint~\cite{3Dxpoint}.
In contrast, in this paper, we design \Design, a fully digital PIM-based DNN accelerator based on single-level memristor devices. \Design removes the necessity of using costly analog/mixed-signal blocks by performing all DNN computations in a digital way, thus providing higher throughput/area.

In digital domain, work in \cite{eckert2018neural} proposed a neural cache architecture which re-purposes caches for parallel in-memory computing. Work in~\cite{li2017drisa} modified DRAM architecture to accelerate DNN inference by supporting matrix multiplication in memory.
In contrast, \Design works on a storage-class memory that can fit the big data. In addition, \Design neuron-to-memory transformation removes the majority of the multiplications involve in DNN and performs non-destructive bitwise operation inside non-volatile memory block without using any sense amplifier. 

\squeezeup
\section{Conclusion}
In this paper, we propose \Design, a fully digital and scalable DNN accelerator. 
\Design framework approximately models all fundamental DNN operations using crossbar memory and associative memory capable of searching nearest distance values. 
We show that the reinterpreted model retains sufficient accuracy of inference quality, and enables the digital-based memory-based computations. 
Our evaluations show that \Design achieves 68.4$\times$, 49.5$\times$ energy efficiency and 48.1$\times$, 10.9$\times$ speedup as compared to ISAAC and PipeLayer while ensuring less than 0.5\% quality loss.



\section*{Acknowledgements}
This work was partially supported by CRISP, one of six centers in JUMP, an SRC program sponsored by DARPA, and also NSF grants \#1730158 and \#1527034.\\

\bibliographystyle{ieeetr}
\bibliography{mybibliography}

\end{document}